# Training Models to Extract Treatment Plans from Clinical Notes Using Contents of Sections with Headings[1]

Ananya Poddar,[*] Bharath Dandala,[*] Murthy Devarakonda[+]

[*]IBM Research, Yorktown Heights, NY, USA

[+]Arizona State University, Tempe, AZ, USA

## Abstract

**Objective:** Using natural language processing (NLP) to find sentences that state treatment plans in a clinical note, would automate plan extraction and would further enable their use in tools that help providers and care managers. However, as in the most NLP tasks on clinical text, creating gold standard to train and test NLP models is tedious and expensive. Fortuitously, sometimes but not always clinical notes contain sections with a heading that identifies the section as a plan. Leveraging contents of such labeled sections as a "noisy" training data, we assessed accuracy of NLP models trained with the data.

**Methods:** We used common variations of plan headings and rule-based heuristics to find plan sections with headings in clinical notes, and we extracted sentences from them and formed a noisy training data of plan sentences. We trained Support Vector Machine (SVM) and Convolutional Neural Network (CNN) models with the data. We measured accuracy of the trained models on the noisy dataset using ten-fold cross validation and separately on a set-aside manually annotated dataset.

**Results:** About 13% of 117,730 clinical notes contained treatment plans sections with recognizable headings in the 1001 longitudinal patient records that were obtained from Cleveland Clinic under an IRB approval. We were able to extract and create a noisy training data of 13,492 plan sentences from the clinical notes. CNN achieved best F measures, 0.91 and 0.97 in the cross-validation and set-aside evaluation experiments respectively. SVM slightly underperformed with F measures of 0.89 and 0.96 in the same experiments.

**Conclusion:** Our study showed that the training supervised learning models using noisy plan sentences was effective in identifying them in all clinical notes. More broadly, sections with informal headings in clinical notes can be a good source for generating effective training data. When trained on noisy-data, both CNN and SVM models seem to be equally effective in sentence classification.

## Introduction

Extracting and succinctly presenting treatment plans, has the potential to improve patient care by reducing the need for foraging a large patient record. While sometimes treatment plans are carefully documented only in a specific section of a clinical note with a heading indicating treatment plan (e.g. "Plan:" or "P:"), it is not always the case. Typically, clinical notes are written using section headings, such as "Assessment:" or "Plan:" (See Figure 1). But such headings are not always present and even when present their surface forms may vary considerably. For example, some notes may use the abbreviated form "P:" (for plan) or synonymous labels such as "Recommendation:" to indicate treatment plans. Our analysis showed only 13% of clinical notes have plan sentences under a commonly used plan heading. Furthermore, plan statements can be found anywhere in a clinical note even when sections with plan headings exist.

---

[1] This work was carried out in 2016 and 2017 while Devarakonda was at IBM Research.

The headings when present can serve as a source of high-precision training data locator because we can use the headings and simple rules in software to extract the sentences within the scope of the heading. These can be used as positive instances to train a model, which in turn can recognize treatment plans occurring anywhere in the same note or in other clinical notes even without the sections with headings. In this study, we explored if this method produces effective training data to train an accurate treatment-plan classifier.

For this study, we used de-identified patient records from Cleveland Clinic (USA), provided under an Institutional Review Board (IRB) approval. From these records, we automatically generated the noisy training data as described later in the article.

We used Support Vector Machines (SVM), and Convolutional Neural Networks (CNN) to model the task of recognizing plan sentences anywhere in a clinical note. The models were trained with the noisy training data, and were tested in two different ways: (1) 10-fold cross validation on the training data; (2) evaluation on a set-aside manually annotated dataset. Conventional, precision, recall, and F measures were used to assess their performance. The learning curve, which represents accuracy improvement with increasing training data, was also obtained.

While text classification is a well-studied problem in general domain NLP as well as in biomedical NLP, our contribution is the demonstration of how noisy training data can be generated from a small percentage of clinical notes written with useful headings can be effective in training models that work well on all clinical notes. To the best of our knowledge, neither using section headings to generate training data nor identifying treatment plans in clinical narratives have been studied before.

## Related Work

Several previous studies had explored automatic training data extraction, the classic approaches were in sentiment analysis of online reviews of movies, restaurants, or products [2] [3]. Since the reviews were typically accompanied by a "star" rating, it was useful as the gold standard to learn sentiment rating from the text.

Distant supervision [4] [5] is another well-known method where existing relations in structured data were used to learn relation extraction models, which were then applied to identify unknown entity relations in a corpus. Structured data (such as vitals, lab results, or medication orders) in a patient record could be a source for gold standard for extracting corresponding insights from clinical notes. However, a recent study [6] reported mixed results in using blood pressure measurements in a patient record as the gold standard for extracting text that indicates hypertension status in clinical notes.

Support Vector Machines (SVMs) [7] [8] have long been used as robust classifiers and therefore, as a sentence classification task, we experimented using SVMs with engineered features, for plan identification. Convolutional Neural Networks (CNN) have been shown to achieve state-of-the art performance in sentence classification in general text [9], which motivated us to experiment with CNNs for the task as well.

## Datasets for Plan Extraction

Datasets used in these experiments were drawn from 1001 longitudinal, de-identified patient records obtained from Cleveland Clinic (USA) with an IRB approval. The patient records had different types of clinical notes depending on the point of contact and the patient care event. We used all unstructured documents, including discharge summaries and admission notes but excluding imaging/lab reports and telephone contact reports, from the patient records. Some patient records contained as few as 3 clinical notes and some others had as many as a few hundred clinical notes. The total number of clinical notes were 117,730. These notes served as a source for the automatically generated as well as manually annotated datasets.

### Automatically Extracting Training Data

Figure 1 shows a part of an anonymized clinical note, highlighting three elements of interest in the automatic training data generation process. The solid box shows the *Assessment & Plan* section, and the dashed and dotted boxes within the section show plan sentences. The dashed boxes show plans with headings, which in this example are "P:" and "Plan:". As mentioned earlier, several variants of these

```
OFFICE VISIT: February 27, 2014
Name: George W. Smith

PCP: Stanley Welby, MD

HPI: Feeling fine.  Bought a home BP device and checks it periodically.  Finds that it varies a lot.  Can be as high as 150/95 or 135/90.  Pretty good about his meds.
…

ASSESSMENT:
1.  HTN – started taking home BPs sporadically.  Running 135-150/90-95.  Admits to not taking his meds consistently.  Have reinforced the importance of controlling his BP due to the cumulative risks for CV events.  I explained how to use the feedback from his BP device to help reinforce the importance of his taking his med regularly.  Patient agrees and says he will start taking his BPs regularly and taking his meds.

P: 1. Check home BPs daily; report repeated BPs over 140/90
   2. Reinforced importance of taking meds consistently
   3. May increase meds if home BPs consistently elevated when he is taking his meds regularly
   4. RTC 3 mo

2. Smoking – talked to patient about the combination of smoking and HTN and his risk of MI, stroke.  Encouraged him to follow his wife's advice about joining the smoking cessation class. Enroll in smoking cessation class

3. Hyperlipidemia – will check LDL. Has been under control in the past.
Plan: Lipid panel
```

*Figure 1. A part of a clinical note from a de-identified patient record. The solid box is an Assessment & Plan section, dashed boxes contain plan sections with headings, and the dotted boxes contain plan sentences without section headings.*

headings can be found in clinical notes such as "Recommendation:" and "Instructions:". The dotted boxes show plan sentences without headers. The method for automatic training data generation is shown in Figure 2. The elements of the method are described below.

**NLP Stack:** As a part of a larger ongoing project, a separate set of algorithms and software have been developed to carry out the basic natural language processing (i.e. tokenization, segmentation, parsing) and clinical concept extraction. The output of this software stack is similar to contemporary packages such as cTAKES [10] and Metamap [11], including named-entity linking of terms in clinical notes to concepts in UMLS [12].

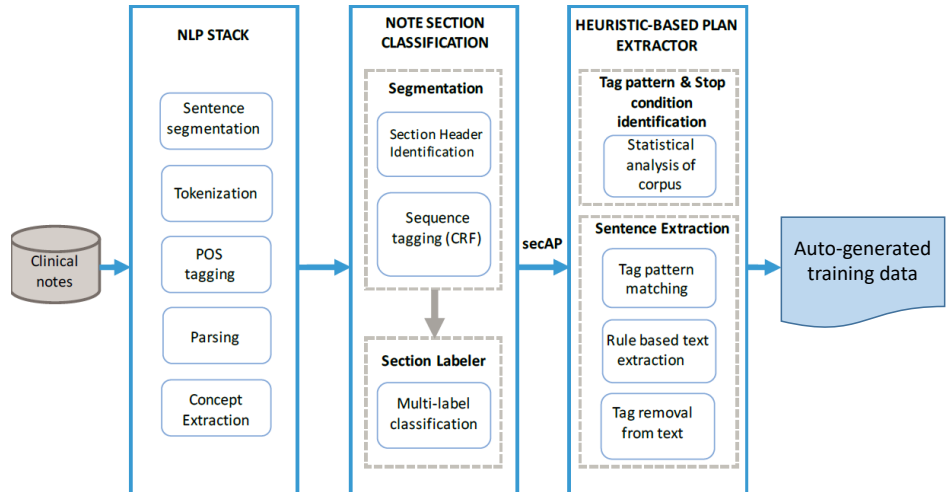

*Figure 2. The automatic training data generation process. The NLP stack provides basic NLP functions and UMLS concept linking. The note section classification segments and labels informal sections in a note. The final stage extracts plan sentences using headings and heuristics.*

**Note Section Classification:** In another part of the ongoing project, software was developed to segment notes into sections and label them in three sequential steps using three different supervised machine learning models. First, a *section-header identification* (supervised learning) model predicts whether a given sentence is the header of a section using several textual and structural features. Subsequently, a *section segmentation* model segments the note into contiguous blocks of sentences using conditional random fields [13]. The section segmenter uses predictions from the section-header identifier as a feature. Finally, *section labeler*, a supervised multi-label classifier assigns section labels to the segmented blocks. We refer to Li, et al [14], for one such implementation. This process identifies Assessment & Plan sections, labels them as *secAP*, in a clinical note such as the one in Figure 1.

From our data, the section labeler identified 46,402 clinical notes as having one or more *SecAP* sections, thus producing a total of 59,014 secAP sections (see Table 2). Note that the section labeler was not trained to identify plan sentences or plan subsections directly. The reasons were twofold: (1) plan sentences with headings are infrequent (only 13% of 59,014 identified secAPs have one or more plan sections with headings) in clinical notes and (2) they are often mixed with assessment statements, especially when headings are not used to denote them as such. Note that the classifier modeled in this paper identifies plan sentences and does not label sections.

**Noisy Training Data Extractor:** We developed a rule-based heuristic plan sentences extractor that works with the secAP sections that were identified by the section labeler. Since secAP contains plans and assessments, we formulated rules that determined the potential start and end of a sequence of plan sentences. We refer to such a contiguous block of plan sentences as a *plan section*.

To develop the patterns and rules, we first manually inspected the corpus to identify frequently occurring plan section headings such as "Plan", "P", "Recommendation", and "Instruction". The goal was to identify headings such that the sentences following them can be identified as plan sentences with high precision. These headings were added to a lexicon, which would trigger the start of the heuristics-based training data extraction process.

As a next step, we performed a statistical analysis of all secAP sections with plan headings to identify a valid *stop-condition* to the process. We observed that the plan sections were typically expressed on a disease basis, starting with a set of assessment statements followed by plan sentences, which were often written as a list. Therefore, the end of such a list, the start of another section indicated by the section heading of a new disease, the start of a new section by the section labeler, or the presence of one or more blank lines worked as a robust stop condition.

To create negative samples for this task, roughly an equal number of (equal to the positive samples) random non-plan sentences in clinical notes were extracted. Note that these sentences came from the potential non-plan part of secAP as well as from other sections of a note. We removed duplicate sentences (which do not help training very much) and (plan) section headings to create the auto-generated training data.

### Manually Annotated Gold standard

A set of 61 clinical notes were randomly selected among Progress Notes and Discharge Summaries from the patient records for the manual annotation task. Five notes were Discharge Summaries and the rest were Progress Notes, reflecting their typical proportion in the dataset. An MD has annotated these notes and reviewed them with an experienced Internal Medicine physician. As in the case of the auto-generated training dataset, the annotated plan sentences were positive examples and an equal number of other sentences were negative samples. Note that this dataset was only used for testing purposes, and not for training the models.

## Methods

We experimented with Support Vector Machines (SVM) [15] and Convolutional Neural Networks (CNN) [9] models for plan sentence classification. The task is a binary classification of sentences, subsequent to sentence detection by the parser in our NLP software stack.

### Support Vector Machines

We used the linear SVM kernel with default hyperparameters. Features for the SVM model were obtained at a sentence level, as it was the unit of classification. The features used were the following:

**Bag-of-words:** A bag of all words in a sentence in a lemmatized form.

**Syntactic n-gram features:** Prior research showed that the use of syntactic parse-based n-gram features instead of traditional n-gram features was effective in classification tasks [16]. Thus, we used a medical domain-adapted English Slot Grammar parser [17] to obtain the dependency based syntactic tree for the sentences. Based on the dependency-parse tree provided by the parser, we obtained possible paths of lengths 2, 3, 4 and included them as features.

**Medical Concepts:** UMLS [12] concepts (Concept Unique Identifiers) in a sentence as features.

**Morpho-syntactic features of the verbs:** We also used an array of features that was derived from the morpho-syntactic properties of verbs in the sentences. The primary properties and distinctions included the position of the verb (i.e. head of a main clause or an auxiliary, which marks tense and aspect) and the tense markings themselves (past, present or future). We considered these features to be important in identifying plan sentences as they are typically expressed using future tense or imperative verbs. (Example: *The patient will be sent for an MRI to further evaluate the knee*). On the other hand, an assessment statement is typically written using past tense verbs. (Example: *The patient was able to go through PT but at a much slower pace*).

**Assertions:** As part of a larger ongoing effort, we developed a supervised learning model that identifies clinical assertions on medical concepts. Specifically, we added negated and hypothetical assertions on clinical concepts as features. We consider these features useful as disease assessments often have negations (Example: *No significant changes since the prior exam*), whereas the treatment plans often contain hypotheticals (Example: *Call back if symptoms persist or worsen*). The assertions and the methodology for identifying them was similar to the i2b2 2010 challenge as reported in Uzuner [18].

**Global Features:** We used section-labels identified using note section classifier (introduced earlier), note type (progress note, discharge note etc.), note category (primary care, test reports, specialty category etc.), and provider type (physician, social worker, registered nurse practitioner etc.) as features.

**Feature Selection:** Instead of using all the features for training our classifier, we used Pointwise Mutual Information (PMI) [19] and Fisher exact test [20] to select features that are likely to have the most impact on classification.

We performed our experiments with SVMs using: (1) only bag-of-word features (SVM BOW) as an SVM baseline and (2) all the above-mentioned features (SVM All Features) that satisfied the feature selection criteria.

## Convolutional Neural Networks

Now popular Convolutional Neural Networks (CNN) utilize layers with convolving filters and can thus exploit local features [21]. Our CNN architectures are based on models proposed by Kim [9]. Table 1 shows the model parameters used in our experiments, and Figure 3 shows the neural network architecture.

Table 1. CNN neural network parameters

| CNN Parameter | Value |
| --- | --- |
| Embedding size | 200 |
| Filter sizes | 2,3,4 |
| Number of filters | 100 (for each filter size) |
| Batch size | 64 |
| Regularization | L2 regularization |
| Dropout probability | 0.5 |
| Dev set | 10% of training data |

**Embedding layer:** Embedding layer maps every word with its corresponding low dimensional feature vector. As an initial experiment, we randomly initialized the word vectors in the embedding layer. Prior research suggests initializing word vectors with those obtained by training an unsupervised neural language model on a large domain-dependent corpus is effective. Thus, we also experimented with two distinct pre-trained word vectors built using word2vec [22] -- (a) learned on PubMed articles [23] and (b) learned on patient records in our dataset. In addition to words in the sentence, the global features identified in the previous section were also used as input to the model. The global features were initialized using random embeddings.

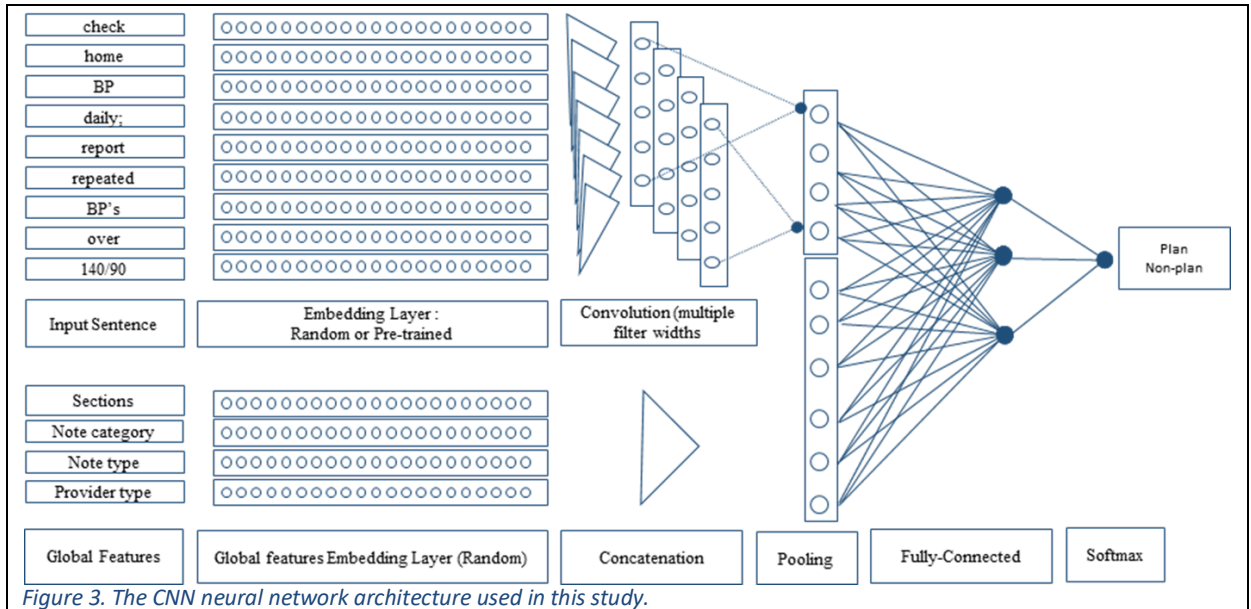

Figure 3. The CNN neural network architecture used in this study.

In our neural network architecture, the embedding layer is followed by a convolution layer of different filter sizes, a 1-max pooling layer, a fully connected feed forward neural network layer, and a softmax classifier as the output layer. For further details of the architecture, we refer to Kim [9]. The global features' embeddings were concatenated with the outputs of max-pool layer from the word-level inputs (see Figure 3). It is worth noting that global features are not at word-level but at sentence-level and they are not sequential like words, thus we did not feed them directly at the word-level input layer.

Table 1 shows the model hyperparameters used during our training. We used 10% of the data as a development set and employed early stopping during the training process.

**Models:** We performed our experiments with CNNs using the following approaches:

- **CNN random**: In this method, we randomly initialized the word vectors in the embedding layer.
- **CNN PUBMED:** In this method, we used pre-trained word vectors from PubMed articles.
- **CNN EHR:** In this method, we used pre-trained word vectors from the unstructured text in our Electronic Health Records dataset.
- **CNN PUBMED/EHR + GLOBAL**: In this method, we additionally used the global features.

# Experiments

The first analysis is to show the percentage of secAP sections with plan heading sections out of the total secAPs in all the patient records. This would inform us of the practical value of the learned model.

Next, we conducted a set of experiments to determine the effectiveness of supervised learning models trained on the auto-generated training data. The first was a 10-fold cross validation on the auto-generated dataset. We calculated standard precision, recall, and F1 measures for the positive class, as well as the micro and macro-averages of F1 scores across the folds, for both classes. We measured the standard deviation across the 10 folds to empirically estimate accuracy variance.

The second experiment was conducted to test the *generalizability* of the models, by measuring accuracy on the manually annotated dataset. The results would indicate broad applicability of the training. Once again, traditional precision, recall, and F1 measures were calculated for the positive class, and micro and macro-average F1 scores for both classes combined, were computed. In this experiment, we included the automatic training data generation (tested on the manually annotated dataset) as a baseline.

The third experiment measured accuracy improvement of the learned models on both datasets as the amount of (auto-generated) training data is incrementally increased, i.e. the learning curve. We measured the plan extraction accuracy under the assumption that only a fraction of the data is available. We ran 10-fold cross-validation experiments using 1%, 2%, …, 100% of the automatically generated data. We also evaluated the model accuracy using manually annotated data for each fraction of training data. The learning curve would indicate if our approach is able to learn from the training data and that the learning is reaching an asymptotic value with the amount of data we acquired through automatic generation.

*Table 2. Dataset description, including the auto-generated training data and the manually annotated gold standard.*

| Description | Data set used for auto-generated training data | Data set used for manually annotated ground truth |
|---|---|---|
| Patient records | 1001 | 1001 |
| Clinical notes selected | 117,730 | 61 |
| Clinical notes containing SecAP sections | 46,402 | 60 |
| Clinical notes containing SecAP sections with one or more sections with plan headings | 6,231 (13%) | 25 |
| Number of SecAP sections | 59,014 | 69 |
| Number of SecAP sections with plan headings | 12,832 (22%) | 57 |
| Number of sentences in sections with plan headings (positive samples) | 13,492 (33%) | 379 (64%) |

# Results

Table 2 shows statistics for the automatically generated training dataset. Only 6,231 out of total 46,402 notes with secAPs (about 13%) in the data set contained plan sections with headings. There are 59,014 secAP sections in all the notes, and out of which only 12,832 (about 22%) contained plan sections with headings. This suggests that using the *training data generation method* only, would result in identifying only a fraction of the plan sentences. However, the plan sections with headings are sufficient enough that we generated sizable training data consisting of 13,492 positive instances. Roughly an equal number of negatives instances (i.e. non-plan sentences) were randomly selected from the rest of the notes.

The statistics for the manually annotated data set are also shown in Table 2. The number of notes with secAP were 60, out of which 25 notes had plan sections with headings. This dataset had 69 secAPs, out of which 27 had plan sections with headings.

## Automatically Generated Dataset Results

Table 3 shows results for the 10-fold experiments on the auto-generated training data. The SVM BOW model achieved F1 of 0.779 for the positive class, and a micro-F1 of 0.866. SVM All Features model, which is trained using all the features, achieved 0.828 F1 for the positive class and micro-F1 of 0.890. SVM All Features model improved the positive-class F1 score by a relative 6.3% compared with SVM-BOW thus demonstrating the importance of additional task-specific features.

CNN Random, in which the embeddings are randomly initialized, achieved slightly lower accuracy (positive class F1 = 0.805 and micro F1 = 0.872) than the SVM model with all features. CNNs using word embeddings from PubMed and our EHR data improved performance. However, adding the global features to CNN lead to performance better than that of SVM All Features. The CNN initialized with our EHR embeddings and using global features (denoted as CNN EHR+Global) achieved the highest performance (positive class F1 = 0.867 and overall micro F1 = 0.909). While SVM All Features did achieve the best precision, CNN EHR+Global outperformed SVM All Features by every other measure (e.g. positive class F1 was better by 4.7%).

*Table 3. Accuracy results for plan sentence classification on the auto-generated ground truth (10-fold cross validation).*

| Method | Plan Sentences (Positive Class) | | | Positive & Negative Classes |
|---|---|---|---|---|
| | Precision | Recall | F1 | Micro F1 |
| **SVM BOW** | 0.851 | 0.719 | 0.779 | 0.866 |
| **SVM All Features** | **0.856** | 0.802 | 0.828 | 0.890 |
| **CNN Random** | 0.803 | 0.807 | 0.805 | 0.872 |
| **CNN PUBMED** | 0.805 | 0.815 | 0.810 | 0.874 |
| **CNN EHR** | 0.809 | 0.837 | 0.823 | 0.882 |
| **CNN PUBMED + GLOBAL** | 0.827 | 0.874 | 0.850 | 0.899 |
| **CNN EHR + GLOBAL** | 0.834 | **0.904** | **0.867** | **0.909** |

While direct comparison is not possible, it is worth noting that Kim also showed [9] that the CNN models outperformed SVM-based methods with domain-specific features.

## Manually Annotated Dataset Results

The accuracy analysis on the manually annotated data set is shown in Table 4. The baseline heuristics-based method (see the *Experiments* section for its description) achieved high precision, but its recall was poor. This is consistent with our prior observations that only a small percentage of plan sections have headings.

*Table 4. Accuracy results for plan sentence classification on the manually annotated dataset.*

| Method | Plan Sentences (Positive Class) | | | Positive & Negative Classes |
|---|---|---|---|---|
| | Precision | Recall | F1 | Micro F1 |
| **Baseline** | **0.955** | 0.227 | 0.367 | 0.487 |
| **SVM BOW** | 0.666 | 0.752 | 0.703 | 0.923 |
| **SVM All Features** | 0.885 | 0.798 | 0.839 | 0.962 |
| **CNN Random** | 0.764 | 0.802 | 0.782 | 0.946 |
| **CNN PUBMED** | 0.819 | 0.814 | 0.817 | 0.955 |
| **CNN EHR** | 0.825 | 0.848 | 0.837 | 0.959 |
| **CNN PUBMED + GLOBAL** | 0.867 | 0.871 | 0.869 | 0.968 |
| **CNN EHR + GLOBAL** | 0.870 | **0.895** | **0.882** | **0.971** |

All SVM and CNN models significantly outperformed the baseline. The general pattern of the results was similar to the auto-generated dataset. SVM-All-Features performed better than SVM BOW. CNN models without the global features have slightly under performed the SVM-All-Features model. The global features provided an additional and distinct boost to the CNN models. The CNN model initialized with EHR embeddings and using the global feature outperformed SVM-All-Features in all respects except precision. The CNN model outperformed the SVM by 5.1% in terms of the positive class F1.

While this dataset is relatively small compared to the generated dataset, since it is manually annotated, these results indicate that the robustness of the models trained with the noisy auto-generated training data.

## Learning Curve

The learning curve, showing the improvement of the positive class F measure as the training data is incrementally added to the SVM All Features model, is presented in Figure 4. The Figure shows that the test accuracy improves continuously with increasing training for cross-validation of the automatically generated dataset and for the manually created dataset. The accuracy reaches an asymptotic value after about 30% of the training data is used in the cross-validation experiment using the auto-generated dataset, while the accuracy steadily improved through the entire range as training data is added when the models were tested on the manually annotated dataset.

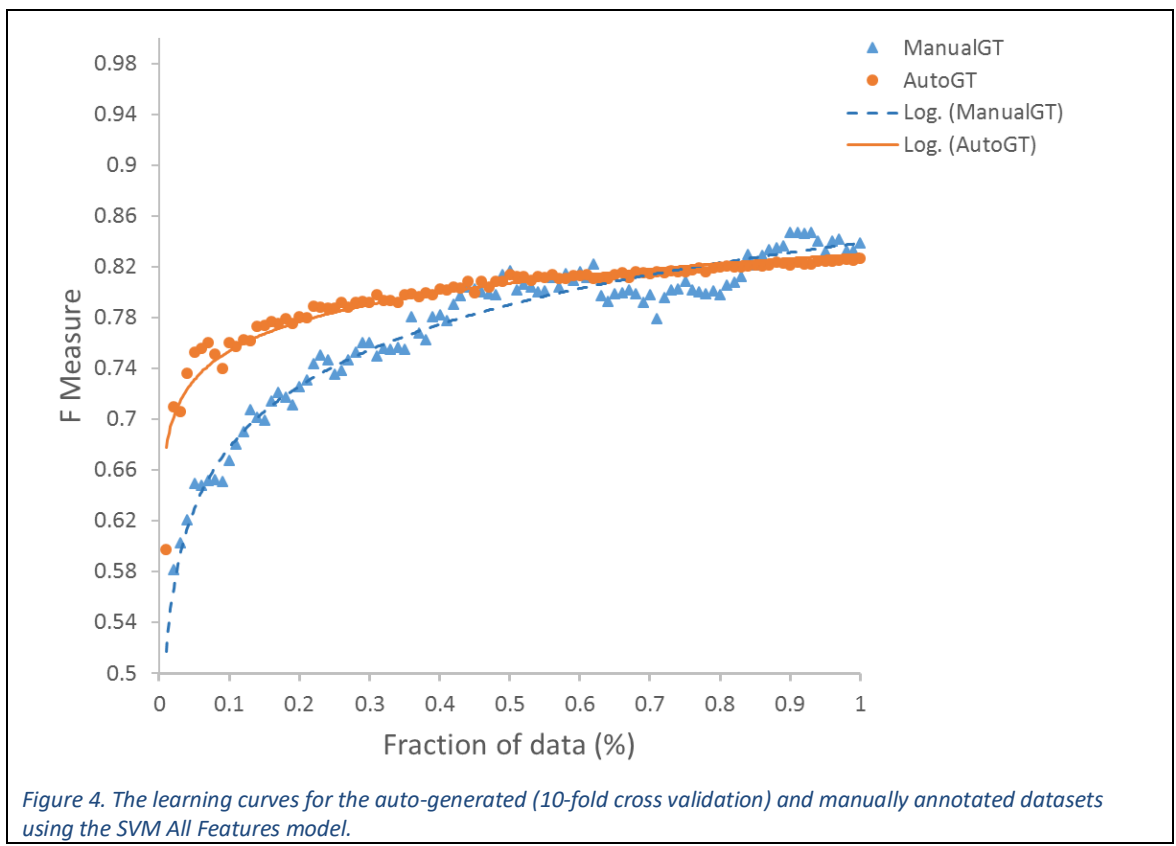

Figure 4. The learning curves for the auto-generated (10-fold cross validation) and manually annotated datasets using the SVM All Features model.

## Discussion

Several points are worth further discussion. Since manual gold standard annotation is expensive and access to clinical notes is constrained by the health care data privacy constraints, automatic training data generation using sections with headings is useful. This study provides a positive data point for successfully generating such training data and for its effectiveness in training well-performing models.

We showed that the sections with headings constitute only about 22% (in 13% of notes) out of the total number of Assessment and Plans sections, and so the models learned using a smaller number of clinical notes can be applied to a larger number of remaining 78% sections (in 87% of notes). To the best of our knowledge such automatic generation of training data and its effective use in training accurate models were not studied previously. Also, to the best of our knowledge, there are no clinical NLP methods that extract plan sentences from clinical notes, and therefore our study presents a novel application of sentence classification to clinical text, using SVMs and CNNs.

In Table 5, we compared accuracy of our models with other (non-clinical) sentence classification results that were reported by Kim [9]. We considered only the binary classification experiments. Our best model's (CNN EHR + Global) performance on the auto-generated dataset was better than three of the four non-clinical sentence classification experiments. Whereas, it's performance on our manually annotated dataset was better than all binary classification experiments reported by Kim. Therefore, the accuracy achieved by our models was on par with or better than the performance of the best models on non-clinical text corpora. Since the datasets and the neural networks are different, caution should be used in interpreting the results.

*Table 5. Comparison with previous non-clinical sentence classification reports.*

| Dataset and Classification (Report Describing the Dataset) | Model (Original Report) | F1 measure for binary classification |
|---|---|---|
| Movie reviews, one sentence per review, positive/negative classification [24] | CNN non-static [9] | 0.815 |
| Product reviews, predict positive/negative reviews [25] | CNN multichannel [9] | 0.850 |
| Opinion polarity detection subtask [26] | CNN static [9] | 0.896 |
| Subjectivity dataset, rating sentences as subjective/objective [27] | Fast-Dropout Neural Networks [28] | 0.936 |
| Auto-generated plan sentences dataset, plan and non-plan sentences classification (this study) | CNN EHR + Global (this study) | 0.909 |
| Manually created plan sentences dataset, plan and non-plan sentences classification (this study) | CNN EHR + Global (this study) | **0.971** |

# Conclusion

In this paper, we described an approach for using sections with headings as a source for automatic training set generation, which can be used to train models to identify treatment plans anywhere in a clinical note. Using available sections with plan headings from clinical notes (about 13% in our patient records dataset), we showed that we can generate sufficient training data to build robust models. The 10-fold cross validation on the auto-generated dataset showed that both SVM and CNN models can accurately classify plan sentences from non-plan sentences. Further accuracy analysis on a manually annotated data set verified the accuracy of these models and thereby, the effectiveness of training with auto-generated data.

In future, we propose to build a fine-grained classification model to further classify the extracted plan sentences into different categories such as medication changes, referrals, contingency plan, lifestyle related goals or a follow-up instruction. Another research direction is to identify *disease specific* plans which have the potential to improve patient care by reducing the need for foraging a large clinical record for critical insights, such as a plan chronology.